\RequirePackage{fix-cm}

\documentclass[smallcondensed]{svjour3}     

\smartqed

\usepackage{graphicx}
\usepackage{amsmath,amsfonts,amssymb,mathrsfs,varwidth}
\usepackage{makeidx}
\usepackage{tikz, rotating,color}
\usepackage{array,booktabs}
\usepackage{algorithm,algorithmicx,algcompatible,algpseudocode}
\usepackage{url}
\usepackage{setspace}

\definecolor{light-gray}{gray}{0.65}
\usepackage[caption=false]{subfig}
\usepackage{verbatim}
\usepackage{placeins}
\usepackage[misc]{ifsym}
\usepackage{doi}
\usepackage{bbding}
\usepackage{hyperref}
\usepackage{microtype}
\usepackage{breakcites}
\usepackage{todonotes}
\usepackage{rotating}
\usepackage{algorithm,algorithmicx,algcompatible,algpseudocode}
\usepackage{wrapfig}
\usepackage[numbers]{natbib}
\usepackage{verbatim}
\usepackage{lineno}
\modulolinenumbers[5]

\journalname{Knowledge and Information Systems}

\begin{document}

\title{Expert-driven Trace Clustering with Instance-level Constraints
}
\author{Pieter De Koninck \and Klaas Nelissen \and Seppe vanden Broucke \and Bart Baesens \and Monique Snoeck \and Jochen De Weerdt}
\institute{P. De Koninck, K. Nelissen, S. vanden Broucke, B. Baesens, M. Snoeck, J. De Weerdt \at
           KU Leuven, Research Center for Management Informatics (LIRIS), Naamsestraat 69, B-3000 Leuven, Belgium \email{pieter.dekoninck@kuleuven.be} \\
           B. Baesens \at
           University of Southampton, Southampton Business School, Southampton, United Kingdom
           }

\date{Received: 06 Sep 2018 / Revised: 06 Jan 2021 / Accepted: 09 Jan 2021}


\maketitle

\begin{abstract}
Within the field of process mining, several different trace clustering approaches exist for partitioning traces or process instances into similar groups. Typically, this partitioning is based on certain patterns or similarity between the traces, or driven by the discovery of a process model for each cluster. The main drawback of these techniques, however, is that their solutions are usually hard to evaluate or justify by domain experts. In this paper, we present two constrained trace clustering techniques that are capable to leverage expert knowledge in the form of instance-level constraints. In an extensive experimental evaluation using two real-life datasets, we show that our novel techniques are indeed capable of producing clustering solutions that are more justifiable without a substantial negative impact on their quality. 
\end{abstract}

\section{Introduction}

Process mining is a research field at the crossroads of data mining and business process management. Its main reason for existence stems from the vast amount of data that is generated in modern information systems, and the desire of organizations to extract meaningful insights from this data. Generally speaking, three subdomains exist within process mining: process discovery, a set of techniques concerned with the elicitation of process models from event data; conformance checking, a set of techniques that aim to quantify the conformance between a certain process model and a certain event log; and process enhancement, approaches that aim to extend existing or discovered process models by using other data attributes such as resource or timing information \citep{VanderAalst2012}.

One of the main challenges in applying techniques from the process discovery subdomain to real-life cases, however, is that the event data corresponding to these cases typically contain highly varied and complex behavioural structures. This leads to a lower quality of the process models which can be discovered. Multiple avenues for mitigating this issue have been proposed, such as focusing on models that hold locally rather than globally \citep{tax2016mining}, applying techniques from the sequential pattern mining domain to extract frequent patterns rather than process models \citep{Mabroukeh2010}, improving the abstraction level of the data \citep{mannhardt2016low} and partitioning the event log into separate clusters, also called trace clustering \citep{Song2009}. 

Trace clustering improves the quality of a process discovery exercise by splitting a highly varied event log into several different clusters of traces, and then discovering a process model for each trace cluster separately. This should decrease the variability of behaviour present in each cluster, and lead to a higher quality of the discovered process models. There have been successful applications of trace clustering techniques in a wide variety of contexts, ranging from e.g. incident management to healthcare \citep{DeWeerdt2013,Delias2015}. 

Nonetheless, trace clustering, like traditional data clustering, is hindered by its unsupervised nature: it is often hard to validate a clustering solution, even for domain experts. This problem has been recognized in \cite{DeKoninck2017Explaining}, in which an approach is proposed to increase understandability of trace clustering solutions by extracting short and accurate explanations as to why a certain trace is included in a certain cluster.
Although explaining cluster solutions to domain experts is a valid approach for enhancing the understandability of trace clustering solutions, it remains a post-processing step. A potentially better approach for improving trace clustering solutions is to directly take an expert's opinion into account while performing the clustering. 

As such, the core contribution of this paper is the proposal and evaluation of two novel types of trace clustering techniques: similarity-driven (or process model agnostic) and process model aware constrained trace clustering techniques. These techniques incorporate expert knowledge, in the form of instance level must-link and cannot-link constraints, into the trace clustering algorithm.
In an experimental evaluation using two real-life datasets, these approaches are shown to lead to clustering solutions that are more in line with the expert's expectations, without substantially diminishing the quality of the clustering solution. 

In light of this objective, the rest of this paper is structured as follows: in Section \ref{sec:potential}, the fields of trace clustering and constrained clustering are described. In Section \ref{sec:alg} their strengths are combined, leading to our proposed approaches. Subsequently, the contribution of our novel approaches is evaluated in Section \ref{sec:eval}. Finally, a conclusion and outlook towards future work are provided in Section \ref{sec:conc}.

\section{Related Work}

\label{sec:potential}
In this section, first, a short overview of trace clustering is provided. Then, the concepts of the constrained clustering fields are described, since they provide an interesting avenue for incorporating expert knowledge. Finally, we combine both aspects to assess the research gap that exists regarding constrained trace clustering: no trace clustering technique has been proposed that can take into account expert knowledge, except for \cite{DeKoninck2017Caise}, in which a full expert-based clustering solution is required. As it is quite unrealistic to expect that experts can give such a complete clustering, this paper proposes two distinct approaches to take into account expert information in the form of constraints while clustering traces. 

\subsection{Classical Trace Clustering}
\label{sec:tctech}
Typically, the starting point of a trace clustering exercise is an event log, which is a set of traces. Each trace is a registered series of events (instantiations of activities), possibly along with extra information on the event, such as the resource that executed the event or time information. A trace clustering is then a partitioning of an event log into different clusters such that each trace is assigned to a single cluster. 

A wide variety of trace clustering techniques exist. Broadly speaking, there are three main categories of trace clustering techniques: those based on direct instance-level similarity, those based on the mapping of traces onto a vector space model, and those based on process model quality.
With regards to direct instance-level similarity, i.e. the direct quantification of the similarity between two traces, an adapted Levenshtein distance could be computed as in \cite{Bose2009}. An alternative set of approaches are those where the behaviour present in each trace is mapped onto a vector space of features \citep{Bose2010,Delias2015}. The third category considers process model quality as an important goal for trace clustering. An approach based on the active incorporation of the process model quality of process models discovered from each cluster has been described in \cite{DeWeerdt2013}. An older approach based on representing the traces in each clusters with Markov Chains has been proposed in \cite{Veiga2010}. 


\subsection{Incorporating Expert Knowledge: Constrained Clustering}
\label{sec:incorpek}
Constrained clustering typically deals with forcing certain instances to be clustered together (must-link or positive constraints), or in separate clusters (cannot-link or negative constraints) \citep{Wagstaff01constrainedkmeans}. A wide array of clustering techniques have been extended to incorporate such constraints, including partitional approaches such as k-means clustering \citep{Wagstaff01constrainedkmeans}, hierarchical clustering \citep{Davidson2005}, model-based clustering with Expectation-Maximization \citep{Law2005}, spectral approaches \citep{Wang2010spectral}, and multi-view clustering \cite{Eaton2014}, among others. Observe that other types of constraints have also been proposed: examples include the use of multi-instance constraints \citep{klein2002instance} or constrained cluster sizes \citep{ZHU2010883}. Constrained clustering has been applied in activity clustering, where the clustering is focused on grouping activities together, rather than partitioning event logs \citep{Wang2016}.

Specific attention is given to the quality of the constraints, as it has been shown that their inclusion can lower the performance of clustering techniques. Davidson and Ravi \cite{Davidson2006} define the \emph{informativeness} of a constraint set as the amount of information contained in the set that the algorithm would not be able to infer on its own. If the objective function or bias of the clustering technique is different from the preference of the constraints, the resulting clustering solution will differ significantly from a situation where no constraints were included. Such a constraint set has a high informativeness. Furthermore, a concept of \emph{coherence} between constraints with regards to a distance function is intuitively described as follows: to have a high coherence, must-link and cannot-link constraints should not contradict each other by connecting multiple points from the same neighbourhood. If points $a$ and $b$ are similar, and so are $c$ and $d$, then it makes little sense to have a must-link constraint between $a$ and $c$ while also having a cannot-link between $b$ and $d$.

\subsection{Constrained Trace Clustering}
\label{sec:twoviews}
Three distinct avenues for the incorporation of expert knowledge in trace clustering are discussed in \cite{DeKoninck2017Caise}: expert seeding, constrained clustering and complete expert pre-clustering. The approach that is subsequently developed in \cite{DeKoninck2017Caise} is based on the latter category: it requires a completely pre-clustered set of traces to be provided by the expert. In this paper, we extend the technique in \cite{DeKoninck2017Caise} by proposing a new algorithm (ConDriTrac) that makes it possible to shift from a full expert solution to a small set of constraints in terms of expert input. Furthermore, adapted versions of process model agnostic techniques presented in \cite{Bose2009} and \cite{Bose2010} that incorporate constraints are  proposed as well, denoted as \emph{Constrained direct} and \emph{Constrained vector-based} techniques below.

\begin{figure}[htb]
\begin{center}
\begin{tabular}{ccc}
    & \textsc{Unsupervised}          & \textsc{Expert-Driven} \\
\rotatebox[origin=c]{90}{\parbox{3cm}{\centering \textsc{Process Model\\ Agnostic}}} 
    & \multicolumn{1}{l|}{\parbox{4cm}{\centering Hierarchical direct distance: e.g. GED \cite{Bose2009}\\ Hierarchical vector-based distance: e.g. MRA, kgram \cite{Bose2010}\\ Spectral vector-based \cite{Delias2015}}}
    & {\parbox{4cm}{\centering Constrained direct*\\ Constrained vector-based*}} \\ \cline{2-3} 
\rotatebox[origin=c]{90}{\parbox{3cm}{\centering \textsc{Process Model\\ Aware}}}
    & \multicolumn{1}{l|}{\parbox{4cm}{\centering Expectation-Maximization\\ Markov Model Clustering \cite{Veiga2010}\\ Active trace clustering \cite{DeWeerdt2013}}} 
    & {\parbox{4cm}{\centering ActSemSup \cite{DeKoninck2017Caise}\\ ConDriTraC* }}           
\end{tabular}
\end{center}
\caption{Classification of trace clustering techniques based on process model awareness and incorporation of expert knowledge. Techniques denoted with a ``*'' are proposed in this paper.}
\label{fig:tcclass}
\end{figure}

Figure \ref{fig:tcclass} illustrates the research gap that is being addressed in this paper. From the discussion in Section \ref{sec:tctech}, it is clear that one of the dimensions for classifying trace clustering techniques is whether or not they are \emph{process model aware}: is the clustering procedure guided by an underlying mined process model representing each cluster? As an alternative, clustering techniques can also be \emph{process model agnostic} in which case the clusterings are determined based on the intrinsic similarities of the traces, thus without the need of a process model. 
Orthogonal to the trace clustering dimension is the aspect of expert supervision. From Figure \ref{fig:tcclass}, it can be seen that all but one trace clustering technique cannot take expert knowledge into account. As such, we specifically address this research gap by proposing both process model-aware as well as process model-agnostic trace clustering techniques that can be guided by an expert in the form of cannot-link and must-link constraints. 


\section{Incorporating Expert Knowledge in Trace Clustering}
\label{sec:alg}
This section introduces our proposed constrained trace clustering techniques. Before outlining the methods themselves, we formally define the necessary background concepts in order for the reader to understand the algorithmic details further in the text.  

\subsection{Preliminaries}
\label{sec:prelim}

An event log is a collection of traces (also called process instances). Each trace contains an identifier, a sequence of events, and optional other attributes. Events can be ordered simply as a sequence, but are typically denoted based on a time stamp.

\begin{definition}[Event]
An event is a tuple $e=(p,\tau,a,x_{1},...,x_{n})$ where $p$ is the identifier of the trace it belongs to, $\tau$ the timestamp, $a$ the activity label, and $x_{1},\dots,x_{n}$ any number of additional attributes. Labeling functions $tid: e \mapsto p$, $time: e \mapsto \tau$, and $act: e \mapsto a$ are included.
\end{definition}

\begin{definition}[Trace]
A trace is a finite sequence of events $t=\langle e_1,\dots,e_{|t|}\rangle$, with $|t|$ the number of events in that trace. The events are sequenced based on their time stamp $\tau$ such that $\forall e_{i}, e_{j} \in t: i < j \rightarrow time(e_{i}) \leq time(e_{j})$. The trace identifier is contained in its events: $\forall e_{i}, e_{j} \in t: tid(e_{i})=tid(e_{j})$. This identifier can be retrieved through the labelling function $tid: t \mapsto tid (e_{1})=p$. Two traces are equal if they have the same identifier: $t=s \Leftrightarrow tid(t)=tid(s)$.
\end{definition}

\begin{definition}[Event Log]
An event log $L$ is a set of traces. $|L|$ denotes its cardinality. 
\end{definition}

Sometimes, it is more useful to group traces that have the same control-flow pattern (disregarding their trace identifiers or timestamp). We refer to such traces as \emph{distinct process instances}. An event log where traces are identified purely based on their sequence of activity labels is called a grouped event log. 

\begin{definition}[Grouped Event Log]
In a grouped event log $G$, traces are considered equal if their sequence of events is equal: $t=s \Leftrightarrow  |t|=|s| \wedge \forall i=1,...,|t|: act(t_{i})=act(s_{i}))$. An element of a grouped event log is called a distinct process instance. The grouped event log $G$ is a multiset of such distinct process instances, with $supp(G)$ denoting the number of distinct traces, or support of the multiset. $|G|$ denotes the cardinality, or its total size. The frequency of a distinct trace is the multiplicity of that trace in the grouped event log.
\end{definition}

\begin{definition}[Trace Clustering]
A trace clustering $C$ is a partition of an event log $L$: a set of nonempty subsets of $L$ such that the union of all clusters is equal to the event log, and none of the clusters overlap: $\bigcup_{A \in C}A = L \wedge \forall A, B \in C: A\cap B \neq \emptyset \rightarrow A = B$.   
\end{definition}

\begin{definition}[Must-link Constraint]
A must-link constraint $ml$ is a relation over an event log $L$ presented as an unordered pair of trace identifiers $\{p,r\}$. A trace clustering $C$ satisfies the constraint if there is a cluster $A$ in $C$ which contains both $p$ and $r$: $\exists A \in C: p \in A \wedge r \in A$.  

\end{definition}

\begin{definition}[Cannot-link Constraint]
A cannot-link constraint $cl$ is a relation over an event log $L$ presented as an unordered pair of trace identifiers $\{p,r\}$. A trace clustering $C$ satisfies the constraint if there is no cluster in $C$ which contains both $p$ and $r$: $\nexists A \in C: p \in A \wedge r \in A$.
 \end{definition}

\begin{definition}[Constraint set]
\label{def:constraintset}
A constraint set $\textit{CS}$ over an event log $L$ is the union of $\textit{ML}$, a set of must-link constraints, and $\textit{CL}$, a set of cannot-link constraints, $\textit{CS} = \textit{ML} \cup \textit{CL}$.
\end{definition}

Over a set of constraints, reasoning can be applied to explicitly include all implied constraints. The resulting set is denoted here as the transitive extension. The combination of a must-link and cannot-link constraint is considered to be \textbf{transitive}, meaning that if `$a$ must be linked to $b$' and `$b$ cannot be linked to $c$', then `$a$ cannot be linked to $c$' either. The same holds for transitivity over two must-link constraints. 
 
 \begin{definition}[Transitive extension]
 
 The transitive extensions $\textit{ML}^{+}$ and $\textit{CL}^{+}$ are defined over the constraints sets $\textit{ML}$ and $\textit{CL}$ containing all implicit must-link and cannot-link constraints respectively.
 
 To formally define these extensions, let us first introduce a boolean function $P(\{x,y\},\textit{ML})$ indicating whether there exists a path of must-link constraints between $x$ and $y$. Please observe that a valid expert input would require that:
 $\forall \{x,y\} \in \textit{CL}: \neg P(\{x,y\},\textit{ML})$. The set of all extended must-link constraints $\textit{ML}^{+}$ is:

\vspace{-6pt}

$$\textit{ML}^{+} = \{ \{x,y\} | P(\{x,y\},\textit{ML})\}$$ 


Similarly, $\textit{CL}^{+}$ includes all implicit cannot-link relationships:

\vspace{-6pt}

$$\textit{CL}^{+} = \{ \{x,y\} | \exists a : \{x,a\} \in \textit{ML}^{+} \wedge \{y,a\} \in \textit{CL} \}$$


Finally, let $\textit{CS}^{+} = \textit{ML}^{+} \cup \textit{CL}^{+}$.

\end{definition}
 
\begin{definition}[Connected traces]
\label{def:connected}
Define the set of all traces which are related to a single trace $t$, given a certain constraint set $\textit{CS}$, including $t$, as the connected traces for $t$ on $\textit{CS}$. It is given by the function $cont: (t,\textit{CS}) \mapsto cont(t,\textit{CS})= \{s\|\{s,t\} \in \textit{CS}\} \cup t$. 
\end{definition}

\begin{definition}[Process model]
A process model $\textit{PM}$ is a diagrammatic representation of a process.
\end{definition}

A multitude of languages exists to model processes using diagrams, ranging from flowcharts and UML activity diagrams to workflow nets and BPMN models. For more information, we refer the interested reader to \cite{DBLP:books/sp/DumasRMR18}. 
\begin{definition}[Process discovery technique]
A process discovery technique $\textit{PD}$ is a function that maps an event log $L$ onto a process model $\textit{PM}$. $\textit{PD}: L \mapsto \textit{PD}(L)= \textit{PM}$.
\end{definition}
 
\begin{definition}[Process model quality metric]
A process model quality metric $m$ is a function which returns a numeric value given an event log $L$ and a process model $\textit{PM}$. $m: (L,\textit{PM}) \mapsto m(L,\textit{PM})$.
\end{definition}

Within the field of process mining, measuring the fit between a process model and an event log is often referred to as conformance checking \cite{rozinat2008conformance}. A wide range of specific metrics have been proposed in the literature, typically focusing on dimensions such as recall (e.g. \cite{van2012replaying}) and precision (e.g. \cite{VandenBroucke2014}). 

\subsection{Constrained Direct and Vector-based Clustering techniques}
In this section, a description is given on how the most prevalent set of trace clustering techniques can be adapted for constraints. A number of approaches rely on the quantification of the similarity between two traces \cite{Bose2009,Bose2010}. A first approach is to define similarity or distance between two traces directly: an example is through calculating the number of deletions, insertions, and substitutions it takes to go from trace 1 to trace 2: this what we denote with \emph{GED}, or Generic Edit Distance. The result is a matrix containing the pairwise distances between each of the traces. Applying a standard clustering algorithm, such as Agglomerative Hierarchical Clustering, results in a clustering solution.
Similarly, in \cite{Bose2010}, approaches are described for defining a vector-space by calculating features using all traces in an event log, and then the distance between traces is represented by the distance between their vectors in the model. These features can be simple: one example are k-grams, activities that appear together (a three-gram is then a set of 3 activities that appear together in some of the traces), or more complex: several featurizations are proposed in \cite{Bose2010} that aim to incorporate common process model characteristics, such as parallelism. The Maximal Repeat Alphabet Feature Set or \emph{MRA} is taken as an example of such featurization. Once the features are chosen, all traces can be mapped onto the vector space model. Given the vector space model and a method of quantifying the distance between vectors (Euclidian, Manhattan, etc.), a matrix with pairwise distances between traces is obtained. On that matrix, a standard data clustering technique can be applied, such as Agglomerative Hierarchical Clustering, resulting in a clustered event log.

Our proposal for making such a technique constraint-aware is simple: given pairwise constraints as input, the goal is to adapt the matrix with pairwise distances to incorporate these must-link and cannot-link constraints. This idea is in line with \cite{Davidson2005}. Under the assumption that the matrix contains distances, the solution is to change the pairwise distances to a very small number when two instances have to be linked, and to a large number when they cannot be linked. If the matrix at hand contains similarities rather than distances, these approaches can be flipped: large similarities are induced for must-links and small similarities for cannot-links. These algorithmic concepts have been implemented as a plugin for ProM 6\footnote{ProM is the leading open-source process mining framework for academicians and practitioners, see: \url{promtools.org}.}, and are publicly available from the package ExpertTraceClustering.

For specific configuration of these numbers, preliminary testing has led to the following setup when performing agglomerative hierarchical clustering with Ward's minimum variance method: the distance between two traces is set to zero when they must be linked, and to the maximum distance present in the original matrix times half the number of traces when they cannot be linked. 

\subsection{ConDriTraC: Process Model Aware Constraint-driven Trace Clustering}
In this section, a novel trace clustering algorithm, \emph{ConDriTraC}, which stands for Constraint-driven Trace Clustering is described. It is designed specifically to be driven by expert knowledge. This algorithm has also been implemented as a plugin for ProM 6, and is publicly available in the same package \emph{ExpertTraceClustering} as the non-process model aware constrained techniques explained above. The technique is based on the approach described in \cite{DeKoninck2017Caise}, where the expert knowledge had the form of a complete expert clustering. Here, the algorithm is adapted to work with constraints as input instead.

In general, the technique consists of three phases: 
\begin{itemize}
\item [] \textbf{Phase 1} An initialization phase, during which the clusters are initialized.
\item [] \textbf{Phase 2} A trace assignment phase, during which traces are assigned to the cluster which leads to the best results, if that best result is sufficiently good.
\item [] \textbf{Phase 3} A resolution phase, during which traces that were not assigned in the previous phase, are either included in an additional separate cluster, or in the best possible existing cluster.
\end{itemize}

\algrenewcommand{\alglinenumber}[1]{\scriptsize#1:}
\begin{spacing}{0.8}
\begin{algorithm}[htb]

\begin{algorithmic}[1]

\scriptsize

\Require $G$ := Grouped Event Log, $k$ := the desired number of clusters, $\textit{CS}$ := a constraint set, $\textit{cvt}$ := cluster value threshold, $tvt$ := trace value threshold; $SeparateBoolean$ := true if unassignable traces should be grouped in a separate cluster;
\Require {\color{gray} Configuration: } $\textit{PD}$:= a process discovery technique, $m$:= a process model quality metric 
\Ensure $C$ := An ordered set of clusters

{\textbf{Phase 1: Initialization}}
\State $C$:=$\emptyset$ 
\State $L$:=$G$  {\color{gray}\% $L$ is set of unclustered traces}
\State Obtain a $k$-bounded maximal clique in $\textit{CL}^{+}$.
\State For each trace $s$ in this clique, add a cluster to $C$ and add $cont(s,\textit{ML}^{+})$ to it. $L$:= $L \setminus  cont(s,\textit{ML}^{+})$.
\State If $\lvert C \lvert < k$, choose random trace $s$ from $L$ and add $cont(s,\textit{ML}^{+})$ to $C$ as new cluster until $\lvert C \lvert = k$. Remove $cont(s,\textit{ML}^{+})$ from $L$.

\vspace{3mm}

{\textbf{Phase 2: Trace assignment}}

\State $U:= \{\}$ {\color{gray}\% List of unassignable traces }
\State Order $L$ by multiplicity
\For{$t \in L$} {\color{gray}\% Loop over the distinct traces which were not assigned to a cluster in Phase 1}
\State  $\textit{bestCluster} := -1$ {\color{gray}\% Temporary value for assignment }

 \State $\textit{bestCMV} := -1; \textit{bestTMV} := -1;$ {\color{gray}\% Temporary values for optimization }
 	\For{$c := (0 \to |C|-1$)} {\color{gray}\% Inspect each possible cluster}
	\If{$\nexists$ $s$ $\in$ $C_{c}$: ${s,t}$ $\in$ $\textit{CL}^{+}$ )} {\color{gray}\% There must not be a cannot-link relation between t and any of the traces already in this cluster} 	\State $\textit{PM} := \textit{PD}(C_{c} \cup cont(t,\textit{ML}^{+}))$ {\color{gray}\% Mine a process model including traces $cont(s,\textit{PM}^{+})$}
    \State $tmv$ := $m(\textit{PM},cont(t,\textit{PM}^{+}))$ {\color{gray}\% Get result of metric on just these traces}  
    \State $cmv$ := $m(\textit{PM},C_{c}$ $\cup$ $cont(t,\textit{PM}^{+}))$ {\color{gray}\% Get result of metric on all traces in cluster c}  
    \If {($tmv >= tvt$) $\wedge$ ($cmv >= cvt$)} {\color{gray}\% Check thresholds}  
    \If {$cmv > \textit{bestCMV} \lor (cmv = \textit{bestCMV} \wedge tmv > \textit{bestTMV})$}
        \State $\textit{bestCMV} := cmv; \textit{bestTMV} := tmv; \textit{bestCluster} := c$  
	\EndIf
 	\EndIf
 	\EndIf
	\EndFor

\If{$\textit{bestCluster} >= 0$} {\color{gray}\% If the traces $cont(t,\textit{ML}^{+})$ could be assigned to a cluster}
 \State $C_{\textit{bestCluster}}$:= $C_{\textit{bestCluster}} \cup cont(t,\textit{ML}^{+})$ {\color{gray}\% Add trace to cluster}
\Else  {\color{gray} \% If the traces $cont(t,\textit{ML})$ could not be assigned to a cluster}
\State $U$:= $U \cup cont(t,\textit{ML}^{+})$ {\color{gray}\% Add traces to unassignable}   
   \EndIf 
 \State $L$:= $L \setminus  cont(t,\textit{ML}^{+})$ {\color{gray}\% Remove traces from log}
 
\EndFor 

\vspace{3mm}

{\textbf{Phase 3: Unassignable resolution}}

\If {$\textit{SeparateBoolean}$} 
\State $C_{n_{b}+1} := U$ {\color{gray}\% Add remaining traces to a new cluster}
\Else 
\State Add each trace to the cluster in $\textit{CS}$ using the same procedure as in phase 2, without checking the thresholds anymore (drop requirements in line 20). Furthermore, the trace and cluster metric values are now calculated without rediscovering a process model each time.
\EndIf
   \State \Return $C$ 

\vspace{2mm}

\normalsize
\caption{ConDriTraC - Constraint-driven Trace Clustering}\label{algorithmConstr}
\end{algorithmic}
\end{algorithm}
\end{spacing}

\textbf{Phase 1: Initialization.} 
The first phase is an initialization phase, as presented in Algorithm \ref{algorithmConstr}. In this phase, a $k$-bounded maximal clique of extended cannot-link constraints is determined. In this case, a clique of cannot-link constraints is a set of traces such that all traces in the set are connected through cannot-link constraints in a pairwise fashion. Such a clique is maximal if no trace can be added without the loss of the clique property. Contrary to finding a largest maximal clique, the search is bounded by the number of clusters $k$ so that this procedure can be stopped once a clique is found with a size equalling $k$. If no such clique can be found, the search will return the largest maximal clique (with its size then being lower than $k$).
Next, each member of the obtained clique is included in a separate cluster, along with all traces that must be linked to it. At that point, if there are less clusters than requested, initialize the remaining clusters randomly from all remaining traces, without violating cannot-link constraints.

\textbf{Phase 2: Trace Assignment.} After the initialization, the set of remaining traces to be clustered will be assigned to the cluster they fit best with. This is done by mining a process model, and calculating the \emph{trace metric value} and \emph{cluster metric value} for each cluster. The \emph{cluster  metric value} is based on the correspondence of all traces to the model, both those that were previously added to the cluster and the ones currently being tested.
The \emph{trace metric value}, on the other hand, corresponds to the result the metric returns based on only the traces that will be added to the mined process model. For each trace $t$, the union of $t$ and the traces it must be linked to is denoted as $cont(t,\textit{ML})$, see Definition \ref{def:connected}. If there exists a cannot-link constraint between at least one of the traces of a cluster, and any trace in $cont(t,\textit{ML}^{+})$, the cluster does not need to be checked. Four situations are possible: (1) the \emph{cluster metric value} is the highest one, in which case the cluster is denoted as the current best; (2) the \emph{cluster metric value} is only equal to the current highest value but the \emph{trace metric value} is higher than the current best, in which case the cluster is also denoted as the current best; (3) the values are above the threshold but lower than the current best found in one of the other clusters, in which case the trace will not be added to the cluster which is currently being tested; or (4) these values are below the provided thresholds, and again the trace will not be added to the cluster which is currently being tested.

After determining the best cluster, the set of instances are added to the best possible cluster. If no best possible cluster exists (because the metric values were below the threshold for each of the clusters), the instances are added to the set of unassignable traces.

\textbf{Phase 3: Unassignable resolution}. In the third phase, any remaining traces which were not assigned to a cluster in Phase 2 will be assigned to a cluster. They are either added to a separate cluster (if \emph{SeparateBoolean} is true), or they are added to the best possible existing cluster. Assigning them to a separate cluster creates a sort of `surplus'-cluster, and should make the process models corresponding with the normal clusters of higher quality. However, observe that cannot-link violations within this surplus-cluster are possible. Therefore, in most cases, it makes more sense not to create this extra cluster. Assigning the traces to the best existing cluster, is done  following the same procedure as in Phase 2, with the two exceptions: on the one hand, the trace value metric and cluster value metric are calculated compared to the process models obtained for each cluster after Phase 2, and no longer rediscovered before each assignment (line 17 is disregarded). On the other hand the thresholds no longer need to be checked (line 20 is disregarded). Because the maximal clique of cannot-links was used to initialize the clustering in Phase 1, there is always at least one cluster to which each trace can be assigned without violating the cannot-link and must-link constraints.

\subsection{Configuration of \emph{ConDriTraC}}
In this subsection, a small discussion is provided on how \emph{ConDriTraC} could be configured. While the choice of the two thresholds and the choice whether or not to separate the not-assignable traces are important decisions, these are case-specific decisions. The algorithm allows for parameter configuration: higher thresholds combined with the separation of traces that do not exceed these thresholds will likely lead to small but high quality clusters and one large surplus-cluster, which may be desirable in some cases but not always.

In terms of the metric chosen as input for the clustering, this depends on the expectation of the underlying process models. A wide array of accuracy and simplicity metrics for discovered process models have been described in the literature (e.g. \citep{DeWeerdt2012}). In general, a weighted metric such as the robust F-score proposed in \cite{de2011robust} might be appropriate, since it provides a balance between fitness and precision.  

A similar argument holds for the process discovery technique one could use. A wide array of techniques exist, and our approach can be combined with most of them. Observe that the chosen technique should be able to discover processes with a decent scalability, since a high number of process models needs to be discovered in certain steps of our algorithm. Whereas in \cite{DeWeerdt2013}, the preference goes to Heuristics Miner \citep{weijters2006process}, \emph{ConDriTraC} relies on the more recent and robust Fodina process discovery technique \citep{vandenBroucke2017}. Alternative options include well performing algorithms like Inductive Miner \citep{Leemans2013} or Split miner \cite{Augusto2018}. Note also that, currently, the focus is on procedural discovery techniques, though an extension towards declarative techniques or mixed-model paradigms \citep{DeSmedt2016} could be considered in future work as well.

\subsection{Theoretical complexity}
The theoretical complexity of the Constrained Direct techniques and Constrained Vector-based techniques is dependent on two steps: first, the calculation of the pairwise similarity (for the direct technique), and the calculation of the features present in the vector space (for the vector-based techniques). The former has quadratic time complexity, while the latter has a linear time complexity \cite{Bose2010}. The second step is applying agglomerative hierarchical clustering, standard complexity of which has been shown to be at least $\mathcal{O}(n^2)$ \cite{Murtagh84}. In this specific context, $n$ is the number of unique traces. Combining the complexity of both steps, we can conclude that the Constrained Direct techniques and Constrained Vector-based techniques will have a quadratic time complexity.

The theoretical complexity of ConDriTraC, on the other hand, is largely dependent on the complexity of the chosen process discovery technique, which have been shown to be non-linear in the number of activities present in the event log \cite{vandenBroucke2017}. Furthermore, the calculation of the evaluation metric, and the clustering structure play an important role as well.

Algorithm \ref{algorithmConstr} lists three distinct phases: initialization of the clusters, trace assignment, and resolution of unassignable traces. With regards to the initialization phase, let $c$ be the number of traces included in at least one cannot-link constraint, $m$ the number of cannot-link constraints, and $k$ the desired number of clusters. The goal is then to obtain a clique with a size equal to $k$. As shown in \cite{chen2006strong}, this can be done in $\mathcal{O}(k \times c^k \times k^2) = \mathcal{O}(c^k \times k^3)$ (lines 1-5). For the second phase, with $n$ referring to the number of unique traces in the event log, and $k$ again the requested number of clusters, each step will discover at most $k$ process models, and calculate at most $2k$ process metrics (lines 11-22). This block is thus completed in $\mathcal{O}(k(\textit{PD}+\textit{ME}))$, where $\mathcal{O}(\textit{PD})$ and $\mathcal{O}(\textit{ME})$ are the complexities of the discovery techniques and the metric evaluation, respectively. This is repeated $n$ times (lines 8-29), resulting in an overall complexity for the second phase of $\mathcal{O}(nk(\textit{PD}+\textit{ME}))$. The final phase reassigns the traces that did not meet the thresholds again (lines 30-34), now without rediscovering a process model each time, so worst case, it will have a complexity of $\mathcal{O}(nk(\textit{ME}))$.

To summarize, the algorithm will run in $\mathcal{O}(c^k \times k^3) + \mathcal{O}(nk(\textit{PD}+\textit{ME})) + \mathcal{O}(nk(\textit{ME}))$, i.e. quadratic in the number $k$ of clusters desired, and given the complexity of the process model discovery and metric evaluation, polynomial in the number of unique traces $n$ and clusters $k$.

\section{Experimental evaluation}
\label{sec:eval}
In this section, we will apply a number of existing trace clustering techniques, and several expert-driven trace clustering techniques, on two datasets. The main objective of the experimental evaluation is to demonstrate that our newly proposed constrained trace clustering techniques are indeed capable of producing more justifiable clustering solutions while at the same time not deteriorating clustering quality. Therefore, the obtained clustering solutions are compared in terms of process model quality and justifiability.

\subsection{Setup}

\textbf{Data sets.} The first dataset, MUNICIP, is a set for which the ground truth is known. This event log is a pre-processed version of the dataset used in the BPI Challenge of 2015, a collection of event data from the permit processes of five Dutch municipalities \cite{BPIC15dataset}. The log contains 5649 traces, of which 2502 are distinct. The dataset contains 29 types of activities. A number of additional statistics is presented in Table \ref{tab:datasetstat}. The five distinct municipalities are considered to be the ``true'' clusters of traces. 

The second data set, TABREAD, is described in \cite{DeKoninck2017Caise}, and contains logged behaviour of participants in a study regarding tablet newspaper reading. The ground truth here stems from a description of cluster structures by a marketing expert, based on information such as the reading moment, length of a session, how focused a reader is, how thoroughly the paper is read, etc. The expert knowledge is not expected to correlate with a trace clustering view over the data. The created event log contains a wide variety of behaviour: out of 2900 reading sessions, there are 2794 distinct variants. This log contains 34 activity types. A number of additional statistics is presented in Table \ref{tab:datasetstat}. Furthermore, a small example is presented in Table \ref{tab:readex}.

\begin{table}[ht]
 \caption{Data set statistics}
\label{tab:datasetstat}

\setlength{\tabcolsep}{8pt}
\scalebox{0.82}{
\begin{tabular}{llllllll}

  \hline
Data set & Traces & Distinct &Act. types&  Average & Min & Max & Average \\
        &  & traces    &            &  trace &  trace &  trace&Act. types\\
 & & & & length & length & length &per trace \\
 \hline
 MUNICIP & 5649 & 2502 & 29 & 16.4 & 3 & 76 & 9.5\\
 TABREAD & 2900 & 2794 & 31 & 20.4 & 2 & 115 & 11.1\\

  \hline\noalign{\smallskip}

\end{tabular}
}
\end{table}

\begin{table}[ht]
 \caption{Example event log of the tablet reading process}
\label{tab:readex}

\setlength{\tabcolsep}{8pt}
\scalebox{0.9}{
\begin{tabular}{lllll}

  \hline
Session & Time & Activity Type &  User \\
  \hline\noalign{\smallskip}
1 & 16-06-2015 08:02 & launch & John Doe\\
1 & 16-06-2015 08:03 & read-page-front & John Doe\\
1 & 16-06-2015 08:03 & read-page-politics & John Doe\\
1 & 16-06-2015 08:04 & scan-page-politics & John Doe\\
1 & 16-06-2015 08:04 & inspect-image-sport & John Doe\\
1 & \ldots & \ldots  & \ldots \\
1 & 16-06-2015 08:24 & quit & John Doe\\
\hline
2 & 16-06-2015 08:32 & launch & Jane Doe\\
2 & \ldots & \ldots  & \ldots \\

   \hline
\end{tabular}
}
\end{table}

\textbf{Expert knowledge.}
The expert knowledge is captured in constraints, which are generated randomly from the ground truth for both data sets, with an equal distribution of must-link and cannot-link constraints. Three sets of constraints are included: 1\%, 5\%, and 10\%, where each set contains a number of constraints equal to said percentage of the number of distinct process instances. The smaller constraint sets are subsets of the larger constraint sets. 

\textbf{Techniques.} All included techniques are listed in Table \ref{tab:clustechsetup}, with an indication of whether they are model aware or not and expert-driven or not. Five classical trace clustering techniques are incorporated for comparison: $ActFreq$ and $\mathit{ActMRA}$ \citep{DeWeerdt2013}, two process model aware techniques, \emph{GED} \citep{Bose2009}, a direct instance-similarity technique, and two vector-space model-based methods, \emph{MRA} \citep{Bose2010} and \emph{3-gram} \citep{Song2009}.
Three clustering approaches are included that take expert knowledge into account but are not model aware: the constrained direct technique \emph{ConGED} and two constrained vector-based, i.e. \emph{ConMRA} and \emph{Con3-gram}. Finally, one novel expert-driven trace clustering approach which is process model aware is included: \emph{ConDriTraC}.

\begin{table}
\caption{Clustering techniques compared in the experimental evaluation}
\centering
\label{tab:clustechsetup}
\scalebox{0.72}{
\begin{tabular}{lllccc}
\hline\noalign{\smallskip}
  Shorthand & Technique & Implementation & Process model & Expert\\
  && (Plugin/package) & aware & driven\\
\noalign{\smallskip}\hline\noalign{\smallskip}
\emph{GED}      & AHC - Generic Edit Distance & GuideTreeMiner (ProM 6)& &\\
\emph{MRA}      & AHC - Maximal Repeat Alphabet & GuideTreeMiner (ProM 6) & & \\
\emph{3-gram}      & AHC - 3-grams & GuideTreeMiner (ProM 6)&& &\\
\emph{ActFreq}      & Frequency-based ActiTraC & ActiTraC (ProM 6)&\Checkmark &\\
\emph{ActMRA}      & Distance-based ActiTraC & ActiTraC (ProM 6) &\Checkmark & \\

\emph{ConGED}      & Constraint-based AHC & own plugin (ProM 6) &  &\Checkmark \\
\emph{ConMRA}      & Constraint-based AHC & own plugin (ProM 6) &  &\Checkmark \\
\emph{Con3-gram}      & Constraint-based AHC & own plugin (ProM 6) &  &\Checkmark \\
\emph{ConDriTraC}      & Constraint-based ActiTraC & own plugin (ProM 6) & \Checkmark &\Checkmark \\

\noalign{\smallskip}\hline
 \multicolumn{6}{l}{\emph{AHC: Agglomerative Hierarchical Clustering}}\\
 \multicolumn{6}{l}{Configuration of \emph{ConDriTraC}: \emph{PD}:= Fodina, \emph{m}:= F1-Score, \emph{SeparateBoolean}:= False}\\
 \multicolumn{5}{l}{For TABREAD: \emph{cvt}:=0.27, \emph{tvt}:=0.27; for MUNICIP: \emph{cvt}:=0.50, \emph{tvt}:=0.25;}\\
 
\noalign{\smallskip}\hline

\end{tabular}
}

\end{table}

\textbf{Metrics.} To evaluate the quality of the clustering solutions, a process model is mined for each cluster, using the Fodina technique \citep{vandenBroucke2017}. The accuracy of each process model discovered per cluster is then measured using the F1-score as proposed in \cite{de2011robust}, where $p_{B}$ is a precision metric and $r_{B}$ is a recall metric: 
\begin{equation*}
 F1_{B} = 2 * \frac{p_{B}*r_{B}}{p_{B}+r_{B}} 
\end{equation*}
In this paper, the recall metric we have chosen is behavioural recall r\textsubscript{b} \cite{Goedertier2009}, and the precision metric we use is etc\textsuperscript{p} \citep{Munoz-Gama2010}. 
Finally, a weighted average F-score metric for the entire clustering solution is then calculated as follows, similar to the approach in \cite{DeWeerdt2013}, where $k$ is the number of clusters in $C$ and $n_{i}$ the number of traces in cluster $i$: 
\begin{equation*} 
F1^{WA}_{C}= \frac{\sum_{i=1}^{k}n_{i}F1_{i}}{\sum_{i=1}^{k}n_{i}}
\end{equation*} 

Furthermore, we can calculate the relative improvement of a technique with expert knowledge ($EK$) compared to the best pure trace clustering technique ($TC$) as follows:

\begin{equation*} 
RI(EK, TC)= \frac{F1^{WA}_{EK}}{F1^{WA}_{TC}}
\end{equation*} 

Three situations might arise: (1) $RI > 1$: in that case, the expert-driven technique creates a solution which is able to combine higher ease-of-interpretation with better results in terms of process model quality; (2) $RI = 1$: the expert-driven technique leads to higher ease-of-interpretation from an expert's point of view without reducing model quality; and (3) $RI <1$: there is a trade-off present between clustering solutions which are justifiable for an expert and the optimal solution in terms of process model quality.

Finally, we propose measuring the justifiability of a solution. As defined in \cite{Martens2011} for classification, justifiability measures the extent to which the results of a classification exercise is in line with existing domain knowledge.
Here, justifiability can be measured as the extent to which an expert's expectations are fulfilled: on the one hand, by calculating the percentage of constraints that are violated by a trace clustering solution. These results are calculated based on the non-propagated versions of the constraint set, to avoid calculating the same violation multiple times. As described in \cite{Davidson2006}, the utility of a constraint set can be measured by its informativeness: the extent to which the constraints add information an unconstrained algorithm is not able to infer on its own. If unconstrained approaches violate a high number of constraints, this can be interpreted as high informativeness.
On the other hand, we can use clustering indices to compare how similar two clustering solutions are. Ideally, a clustering solution should be closely related with the ground truth. For this purpose, the Jaccard Index \cite{ben2001stability} is used, which is a measure for the overlap between two clusterings: if n11 is the number of pairs of items that are clustered together in clustering a and b, n10 is the number of pairs of items that are clustered together in a but not in b, and n01 vice versa, then the Jaccard Index of a and b is:
\begin{equation*}
JI(a,b)= \frac{n11}{n11+n10+n01}
\label{eq:JI}
\end{equation*}

\subsection{Results MUNICIP: 5 municipalities}

\begin{figure}
\centering
  \includegraphics[width=0.9\textwidth]{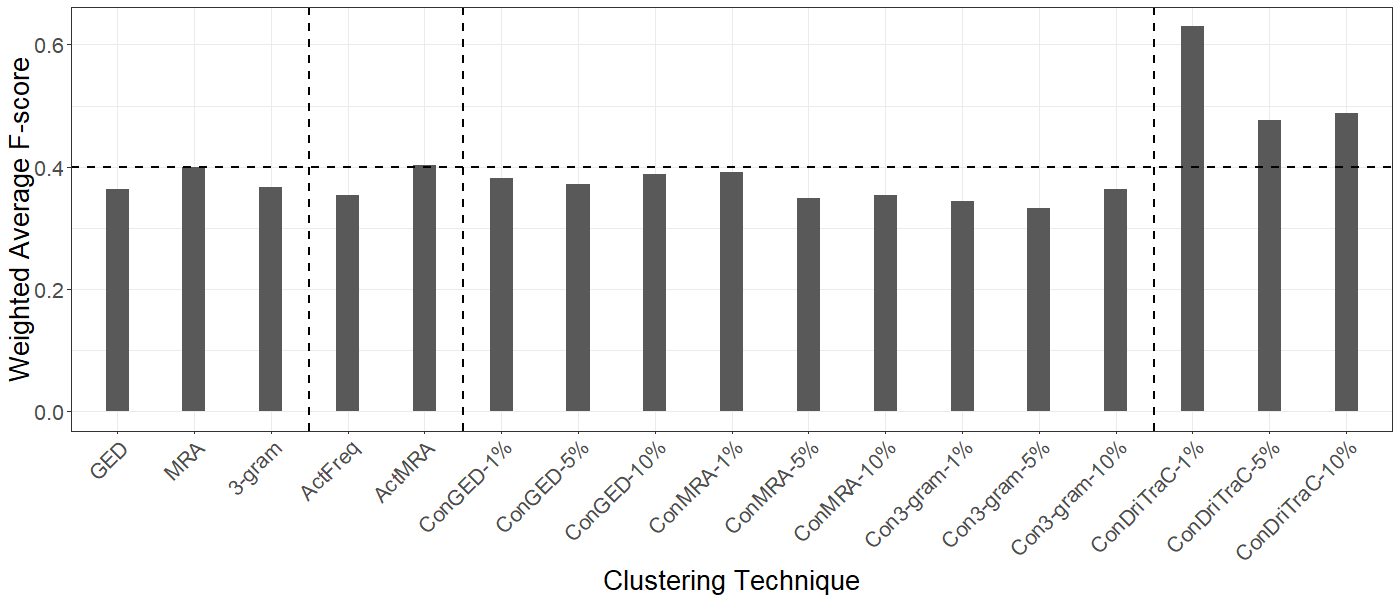}
  \caption{Weighted Average F1-score results on MUNICIP. The horizontal dashed line indicates the baseline F-score without applying any clustering to the data.}
  \label{fig:f1scoreBPIC15}
\end{figure}

The results in terms of F1-score are visualised in Figure  \ref{fig:f1scoreBPIC15}. A number of observations can be made from this figure. First, observe that most F1-scores are rather low. This is due to the fact that dividing 2502 distinct process instances in 5 clusters still leads to rather large clusters. Therefore, the precision of the process models corresponding with these clusters is rather low. Secondly, the horizontal line on Figure \ref{fig:f1scoreBPIC15} represents the F-score of the unclustered event log, which is just around 0.39. Given the results of the other techniques, we see that a surprising number of trace clustering techniques do not improve compared to an unclustered event log. Most notable are the process-aware techniques \emph{ActFreq} and \emph{ActMRA}, who optimize for fitness over precision. 
Of all techniques, the only technique to significantly outperform the others is the constrained process model aware technique, \emph{ConDriTraC}. When given the lowest percentage of constraints, it attains a weighted average F1-score of 0.63, the configurations with the 5 and 10\% constraint set reach 0.48 and 0.49 respectively.
In terms of relative improvement, the constrained direct (\emph{ConGED}) and vector-based (\emph{ConMRA, Con3-gram}) all score worse than the best pure model-driven technique (\emph{ActMRA}), with RI $<$ 1. The process model aware techniques, \emph{ConDriTraC-1\%, ConDriTraC-5\%}, and \emph{ConDriTraC-10\%}, attain relative improvements of 1.56, 1.18, and 1.21, respectively.

\begin{figure}
\centering
  \includegraphics[width=\textwidth]{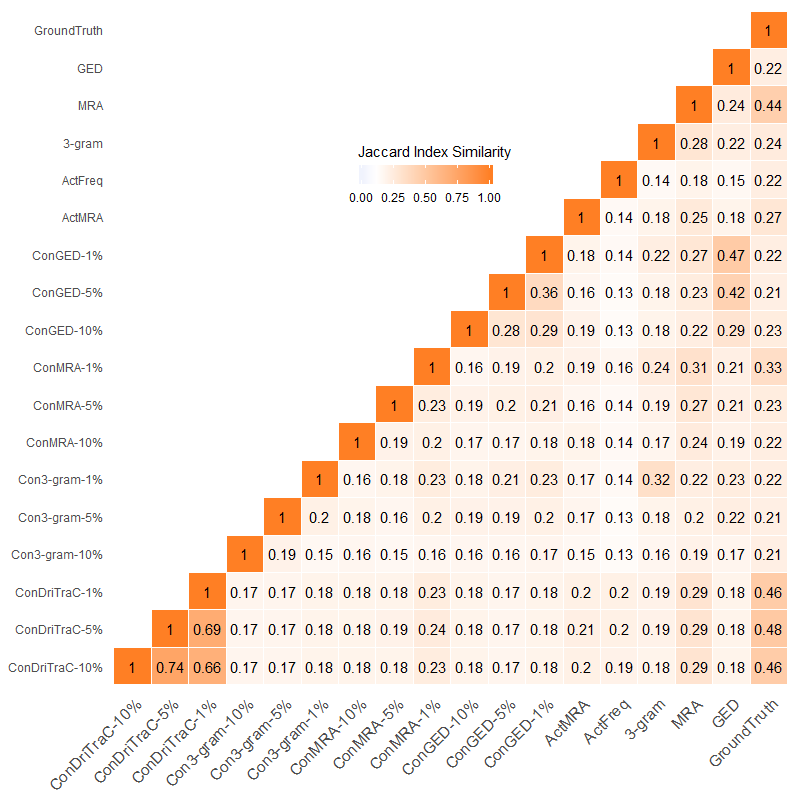}
  \caption{HeatMap representing the Pairwise similarity of the clustering results of each of the clustering techniques, measured using the Jaccard Index on MUNICIP}
  \label{fig:HeatMapJIBPIC15}
\end{figure}

\begin{figure}[htb]
\centering
  \includegraphics[width=0.85\textwidth]{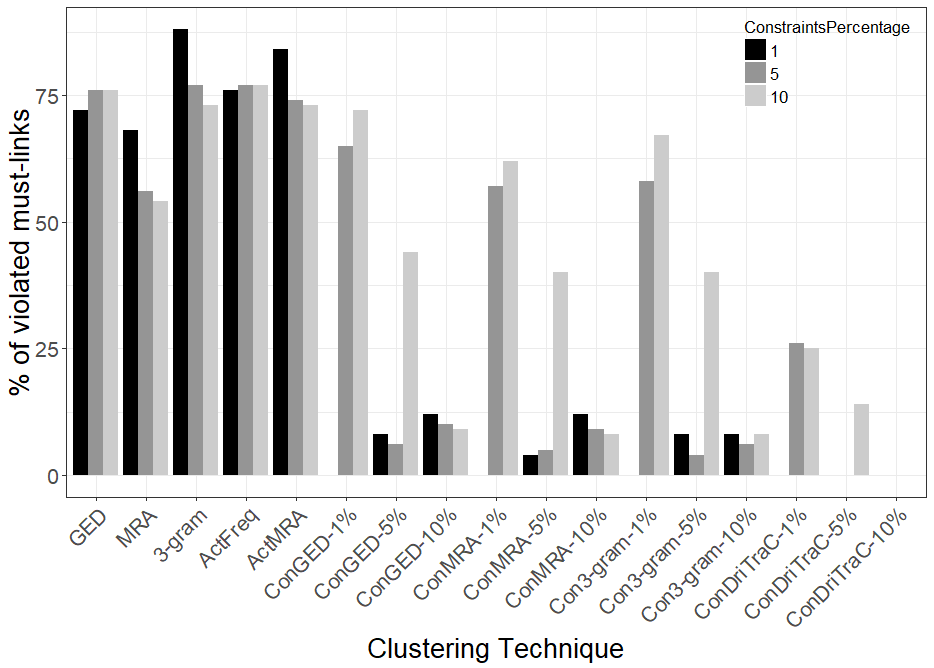}
  \caption{Average percentage of violated must-link constraints for MUNICIP}~\label{fig:violmlBIPC15}
\end{figure}

\begin{figure}[htb]
\centering
  \includegraphics[width=0.85\textwidth]{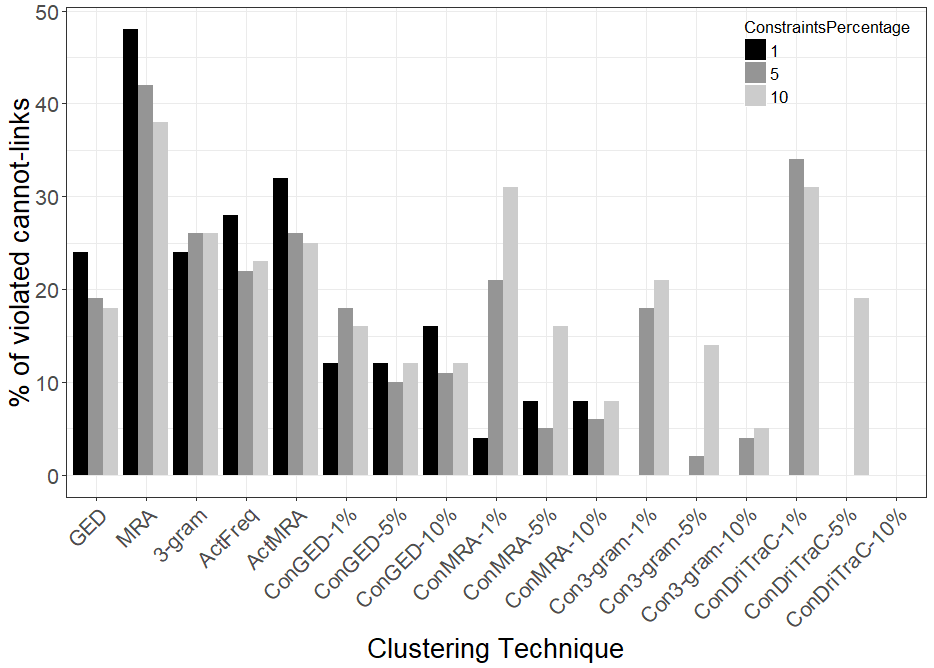}
  \caption{Average percentage of violated cannot-link constraints for MUNICIP}~\label{fig:violclBPIC15}
\end{figure}

Next, Figures \ref{fig:violmlBIPC15} and \ref{fig:violclBPIC15} present the percentage of violated must-link and cannot-link constraints respectively, for each of the techniques and each of the percentage of constraints. 

A couple of remarks can be made: first, observe that \emph{ConDriTraC-10\%} is the only technique that doesn't violate any of the constraint sets. \emph{ConDriTraC-1\%and ConDriTraC-5\%}, and the other constrained clustering techniques, violate a fraction of the constraints they were not provided with. The non-process model aware techniques, \emph{ConGED, ConMRA} and \emph{Con3-gram}, also violate some of the constraints they were given, with the exception of \emph{Con3-gram-1\%}.

Finally, Figure \ref{fig:HeatMapJIBPIC15} represents the similarity of the clustered event logs obtained by each of the clustering techniques. Most interesting is the similarity of the results to the ground truth. Here, we see that the process model aware expert-driven trace clustering technique \emph{ConDriTraC} performs best, closely followed by the unconstrained \emph{MRA}. Of the constrained techniques that are not model aware, \emph{ConMRA} performs best, though its results are closer to the ground truth with less constraints. Finally, observe the high similarity between \emph{GED} and \emph{ConGED}, which decreases when more constraints are added, in line with expectations. Furthermore, the high similarity among the results of \emph{ConDriTraC} based on different constraint sets is noticeable. 

\subsection{Results TABREAD: tablet newspaper reading}

\begin{figure}[htb]
  \centering
  \includegraphics[width=\textwidth]{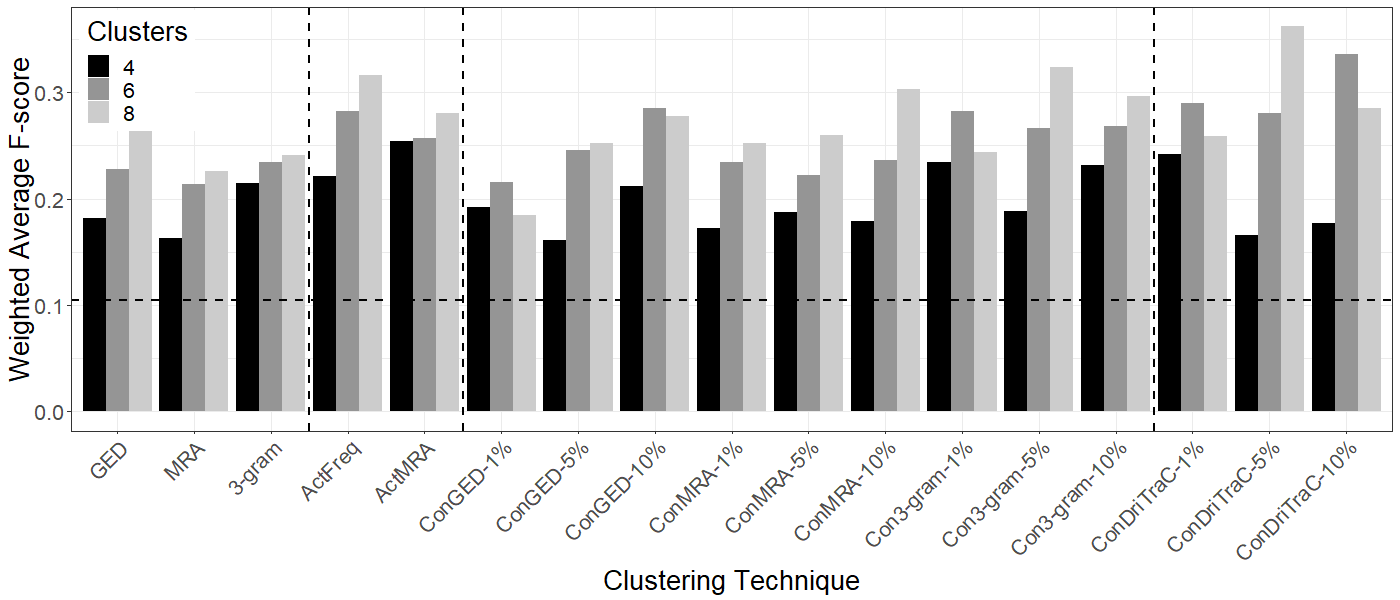}
  \caption{Weighted Average F1-score results for different clustering techniques and number of clusters on TABREAD}\label{fig:f1score}
\end{figure}

The results in terms of F1-score are visualised in Figure \ref{fig:f1score}. Similar to the results on MUNICIP, the F1-scores remain rather low. The reason is still similar: dividing 2794 distinct process instances in 4 to 8 clusters still leads to rather large clusters. Therefore, the precision of the process models corresponding with these clusters is still rather low. Nonetheless, there is a clear improvement when clustering the log on this dataset: compared to the unclustered event log, represented by the horizontal line (F-score of 0.105), all settings lead to increased F1 scores. Secondly, observe that in general, more clusters leads to higher scores, in line with expectations. For the non expert-driven trace clustering techniques (\emph{GED, MRA, 3-gram, ActFreq} and \emph{ActMRA}), the latter two outperform the former three, in line with the findings of \cite{DeWeerdt2013}. Finally, for the expert-driven techniques \emph{ConDriTraC-5\%} performs best for 8 clusters, with \emph{ConDriTraC-10\%} performing best for 6 clusters. For 4 clusters, the best result is obtained by \emph{ActMRA}.

\begin{table}[htb]
\caption{Relative improvement of the constrained process model aware clustering solutions compared to the unconstrained process model aware solutions on the TABREAD dataset}
\label{tab:RI}
\begin{tabular}{lccc}
\hline\noalign{\smallskip}
    & 4 clusters & 6 clusters & 8 clusters\\
    \noalign{\smallskip}\hline
    $RI(\textit{ConDriTraC-1\%},$ \emph{ActFreq} $)$ & 1.10 & 1.03 & 0.82\\
    $RI(\textit{ConDriTraC-1\%},$ \emph{ActMRA} $)$ & 0.95 & 1.13 & 0.92\\
    $RI(\textit{ConDriTraC-5\%},$ \emph{ActFreq} $)$ & 0.75 & 0.99 & 1.15\\
    $RI(\textit{ConDriTraC-5\%},$ \emph{ActMRA} $)$ & 0.65 & 1.09 & 1.29\\
    $RI(\textit{ConDriTraC-10\%},$ \emph{ActFreq} $)$ & 0.80 & 1.19 & 0.90\\
    $RI(\textit{ConDriTraC-10\%},$ \emph{ActMRA} $)$ & 0.69 & 1.31 & 1.02\\
\noalign{\smallskip}\hline
\end{tabular}
\end{table}

For a closer comparison of the results of the expert-driven techniques, Table \ref{tab:RI} contains the relative improvement of the constrained process model aware techniques compared to the unconstrained process model aware trace clustering technique (\emph{ActFreq} and \emph{ActMRA}). For 4 clusters, only \emph{ConDriTraC-1\%} outperforms the non-constrained techniques. For 6 clusters, all constrained solutions perform better or on par with the unconstrained process model aware techniques. Finally, for 8 clusters, the results are split evenly over the unconstrained and constrained solutions. 

\begin{figure}[htb]
\centering
  \includegraphics[width=0.85\textwidth]{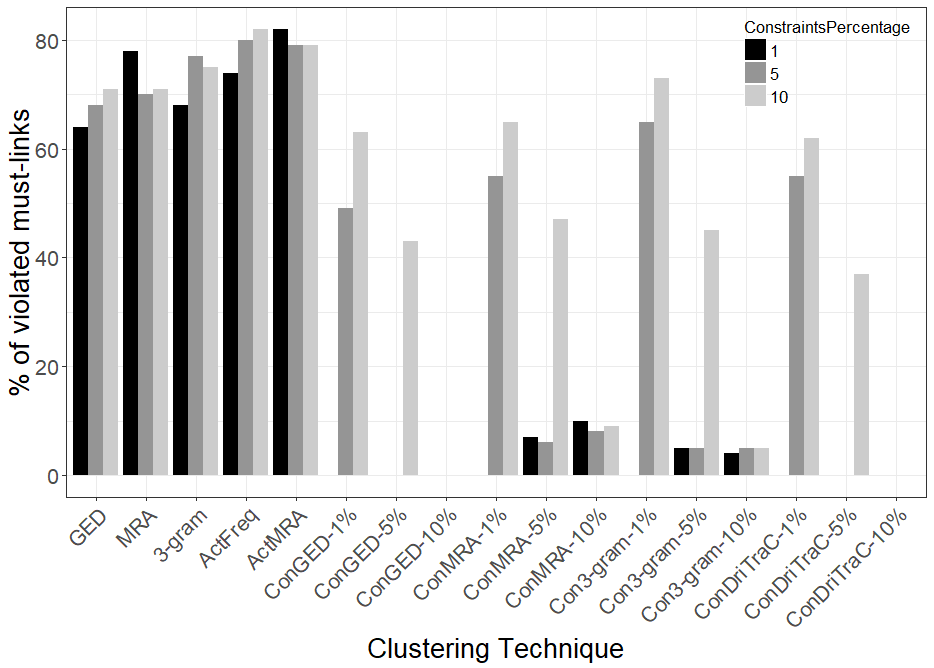}
  \caption{Percentage of violated must-link constraints averaged across 4, 6 and 8 clusters on TABREAD}~\label{fig:violml}
\end{figure}

\begin{figure}[htb]
\centering
  \includegraphics[width=0.85\textwidth]{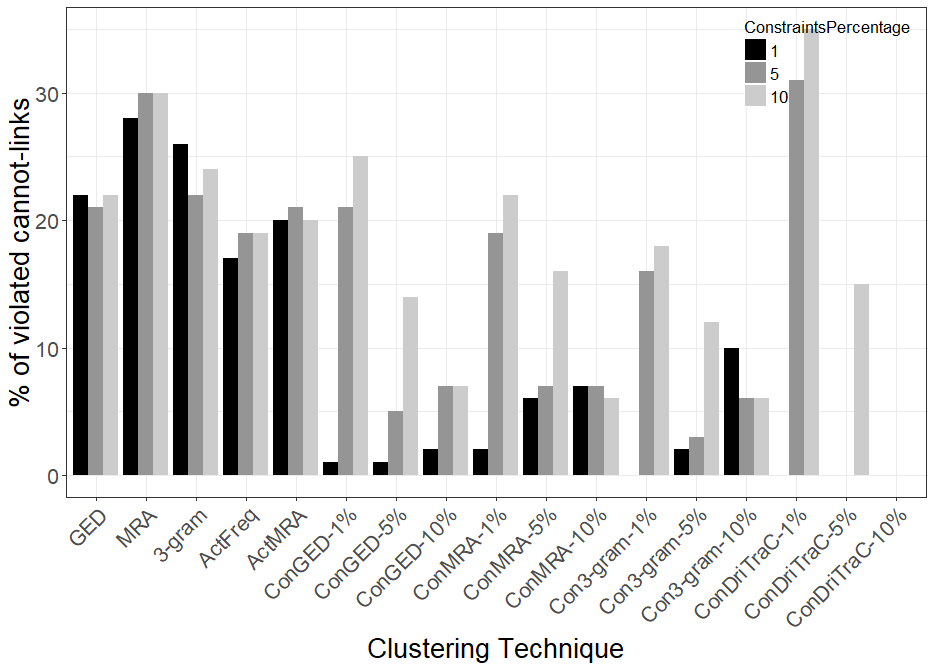}
  \caption{Percentage of violated cannot-link constraints averaged across 4, 6 and 8 clusters on TABREAD}~\label{fig:violcl}
\end{figure}

Next, Figures \ref{fig:violml} and \ref{fig:violcl} present the percentage of violated must-link and cannot-link constraints respectively, for each of the techniques and each percentage level of constraints. The results are averaged across 4, 6 and 8 clusters. These results confirm the findings from the previous section regarding MUNICIP. Observe that the must-link constraints have a higher informativeness than the cannot-link constraints: a higher proportion of must-link constraints is violated by unconstrained techniques compared to cannot-link constraints. This makes sense, especially for higher numbers of clusters: if there are 8 clusters to which one can assign traces, it is less likely that two traces would be assigned to the same cluster by a non-constrained clustering technique, making cannot-link violations less likely, and must-link violations more likely. This is represented in the data: most unconstrained clustering solutions violate between 50 and 80\% of all must-link constraints, whereas they tend to violate 15 to 30\% of cannot-link constraints. 

\begin{figure}[htb]
  \includegraphics[width=\textwidth]{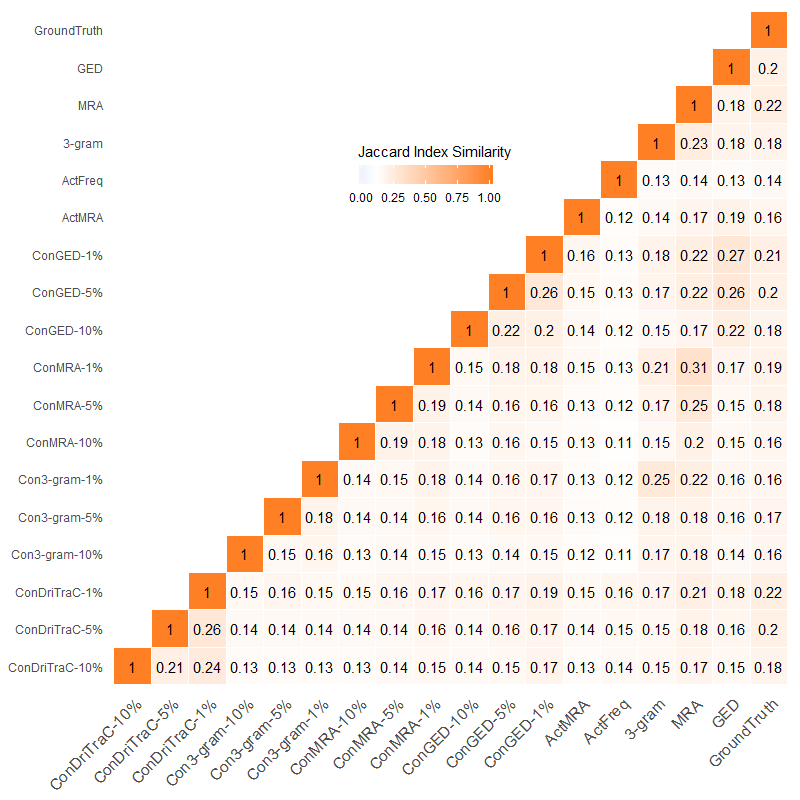}
  \caption{HeatMap representing the Pairwise similarity of the clustering results of each of the clustering techniques, measured using the Jaccard Index, averaged over the cluster numbers, on TABREAD}
  \label{fig:HeatMapJIDSCat}
\end{figure}

Finally, the extent to which the results found are in line with the expectations of the expert (the ground truth) is visualized in Figure \ref{fig:HeatMapJIDSCat}. It is clear that none of the techniques approach the expert's knowledge exceptionally well, with \emph{MRA} and \emph{ConDriTraC-1\%} performing best with a similarity to the ground truth of 0.22. This is especially noticeable when compared to the highest similarity to ground truth on MUNICIP (the previous dataset), where a 0.48 similarity was reported for \emph{ConDriTraC-5\%}. This leads us to believe that on MUNICIP, there is some merit to the ground truth (the 5 distinct municipalities), whereas on TABREAD, the ground truth (a customer profiling provided by a marketeer) is less in line with a trace clustering view of the data.

\subsection{Discussion}
The experimental evaluations of each of the two datasets show that overall, our techniques are indeed capable of discovering trace clustering solutions that are in line with expert input (i.e. not violating constraints). Furthermore, we show that the constrained clustering techniques can maintain a comparable clustering quality, as measured by an aggregated F1-score of all underlying process models of a clustering. For the MUNICIP dataset and for particular settings of the ConDriTrac algorithm for the TABREAD dataset, the inclusion of expert knowledge even improves the clustering quality. This is due to the fact that trace clustering techniques in general cannot guarantee optimality due to computational complexity of the problem. 
As for limitations, it is important to point out that although the findings for both datasets are congruent, generalization towards other datasets is not proven. Moreover, our analysis shows that making a comparison with a ground truth is challenging. On the one hand, for MUNICIP, the trace clustering techniques might be picking up on some other information in the traces rather than whether they were executed within a particular municipality. For TABREAD, the ground truth was based on an extrapolation of reading profiles as defined by an expert. In the latter case, these reading profiles corresponded less to a process-based view on newspaper reading.

\section{Conclusion}
\label{sec:conc}
In a situation where an expert has a preconceived notion of what a clustering should look like, it is unlikely that a trace clustering algorithm will lead to clusters which are in line with his or her expectations. This paper proposes \emph{expert-driven trace clustering} techniques that balance improvement in terms of trace clustering quality with the challenge of making clusters more justifiable for the expert. In an experimental evaluation, we have shown that existing trace clustering techniques lead to solutions that will violate the expert's expectations. Our proposed constrained trace clustering techniques successfully combine the strength of existing trace clustering techniques and constrained clustering approaches.
Furthermore, this papers presents multiple benefits for practitioners: first, all proposed techniques are publicly available. Secondly, we have shown, both intuitively and in our experiments, how must-link constraints are more powerful than cannot-link constraints. This entails that it is more useful to invest time into extracting must-link constraints than cannot-link constraints in a practical setting. Finally, our approach for evaluating trace clustering solutions can be a guideline for deciding which solution to choose, whether it is based on process model quality, or by evaluating the usefulness of the constraint set.

Nonetheless, interesting avenues for future research remain. First, our approach could be applied to more data sets, both real life and artificial. Secondly, the usefulness of other types of constraints, such as constraints based on case-specific data, is a topic for further investigation. Finally, active constraint selection, in which the user is asked to provide answers to constraint queries during the clustering process, could lead to intriguing insights as well.

\section*{Acknowledgements}
This research has been financed in part by the EC H2020 MSCA RISE NeEDS Project (Grant agreement ID: 822214)

\vspace{10pt}

\noindent This is a preprint of an article published in \emph{Knowledge and Information Systems}. The final authenticated version is
available online at: \url{https://doi.org/10.1007/s10115-021-01548-6}.

\vspace{10pt}

\noindent Please cite as follows: 

\begin{description}
\item \noindent De Koninck, P., Nelissen, K., vanden Broucke, S. et al. Expert-driven trace clustering with instance-level constraints. \emph{Knowl Inf Syst} \textbf{63}, 1197–1220 (2021). \url{https://doi.org/10.1007/s10115-021-01548-6}. 
\end{description}

\FloatBarrier

\bibliography{EDTC-bib}

\bibliographystyle{spbasic}      

%
%
\section*{Author Biographies}
\leavevmode

\vbox{%
\begin{wrapfigure}{l}{80pt}
{\vspace*{-20pt}\fbox{\includegraphics[width=3cm]{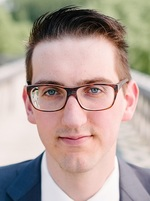}}\vspace*{100pt}}%
\end{wrapfigure}
\noindent\small 
{\bf Pieter De Koninck} obtained his PhD in Business Economics at KU Leuven in 2019. The topic of his PhD was Advanced Clustering Techniques for Business Process Execution Traces. Heretofore, he obtained his Msc in Business Engineering at KU Leuven in 2015. He is currently working as a data scientist at Boltzmann, a Ghent-based AI consultancy firm. \vadjust{\vspace{80pt}}}

\vbox{%
\begin{wrapfigure}{l}{80pt}
{\vspace*{-20pt}\fbox{\includegraphics[width=3cm]{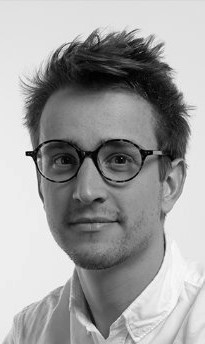}}\vspace*{100pt}}%
\end{wrapfigure}
\noindent\small 
{\bf Klaas Nelissen} graduated with a PhD in Business Economics: Information Systems Engineering from KU Leuven in 2019. The topic of his PhD was “Essays on Meauring User Engagement Online News”. His research interests include measuring user engagement, the correlation of short- and long-term measures, and the link between implicit and explicit measurements of the quality of user experience. He is currently working as a postdoctoral researcher at KU Leuven (Belgium) in the group of biomedical sciences. Klaas is also very interested in technology entrepreneurship and hopes to be able to focus on research valorization in the near future.\vadjust{\vspace{40pt}}}

\newpage

\vbox{%
\begin{wrapfigure}{l}{80pt}
{\vspace*{-20pt}\fbox{\includegraphics[width=3cm]{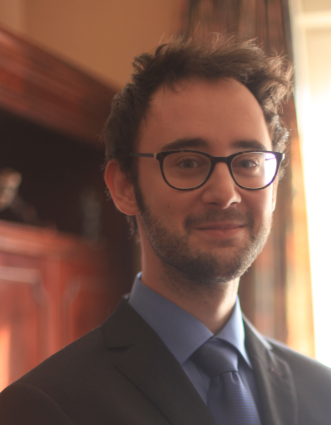}}\vspace*{100pt}}%
\end{wrapfigure}
\noindent\small 
{\bf Seppe vanden Broucke} received a PhD in Applied Economics at KU Leuven, Belgium in 2014. Currently, Seppe is working as an assistant professor at the department of Business Informatics at UGent (Belgium) and is a lecturer at KU Leuven (Belgium). Seppe's research interests include business analytics, machine learning, process management, and process mining. His work has been published in well-known international journals and presented at top conferences. \vadjust{\vspace{80pt}}}

\vbox{%
\begin{wrapfigure}{l}{80pt}
{\vspace*{-20pt}\fbox{\includegraphics[width=3cm]{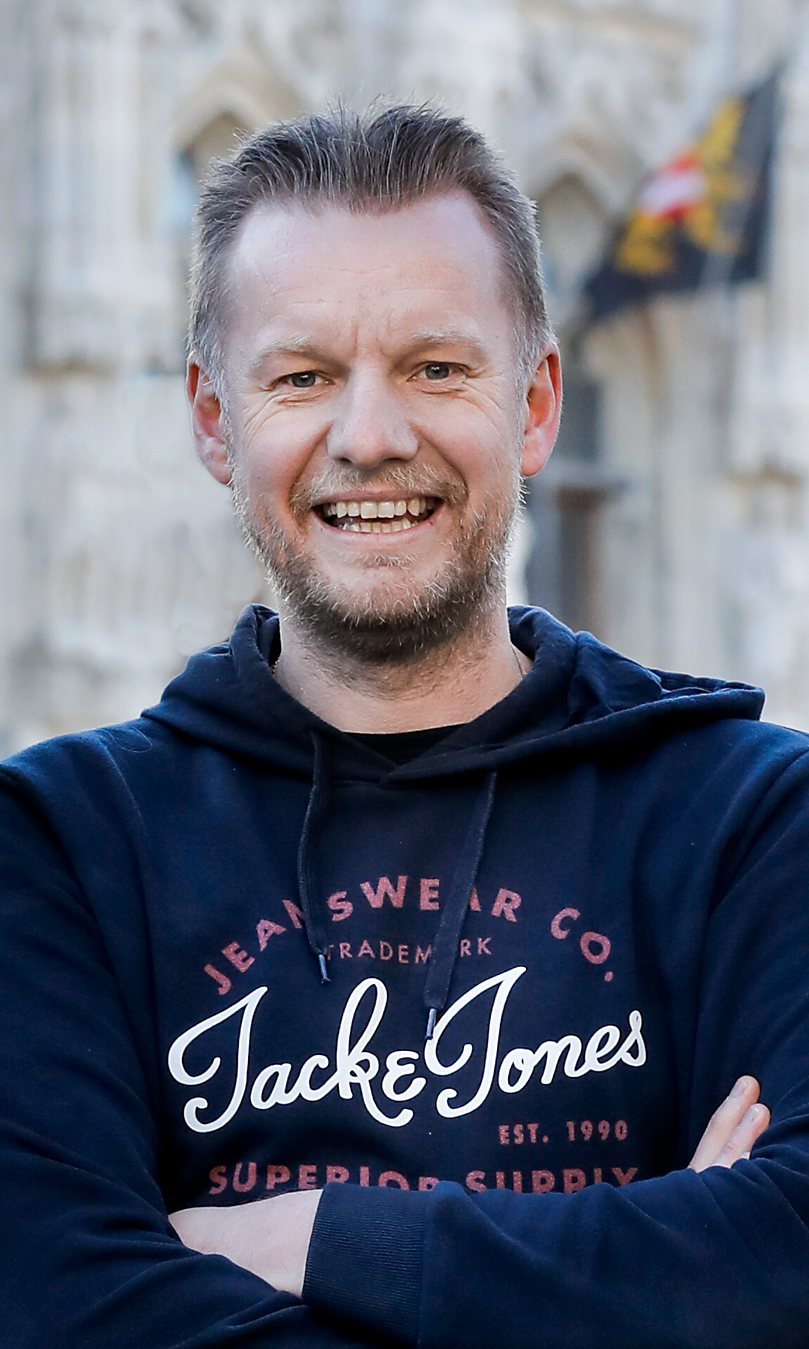}}\vspace*{100pt}}%
\end{wrapfigure}
\noindent\small 
{\bf Bart Baesens} is a professor of Big Data and Analytics at KU Leuven (Belgium), and a lecturer at the University of Southampton (United Kingdom).  He has done extensive research on big data and analytics, credit risk modeling, fraud detection, and marketing analytics.  He co-authored more than 250 scientific papers and 10 books some of which have been translated into Chinese, Japanese, Korean, Russian and Kazakh, and sold more than 30,000 copies of these books world-wide.  Bart received the OR Society’s Goodeve medal for best JORS paper in 2016 and the EURO 2014 and EURO 2017 award for best EJOR paper.  His research is summarized at www.dataminingapps.com.  He also regularly tutors, advises and provides consulting support to international firms with respect to their analytics and credit risk management strategy.\vadjust{\vspace{50pt}}}

\vbox{%
\begin{wrapfigure}{l}{80pt}
{\vspace*{-20pt}\fbox{\includegraphics[width=3cm]{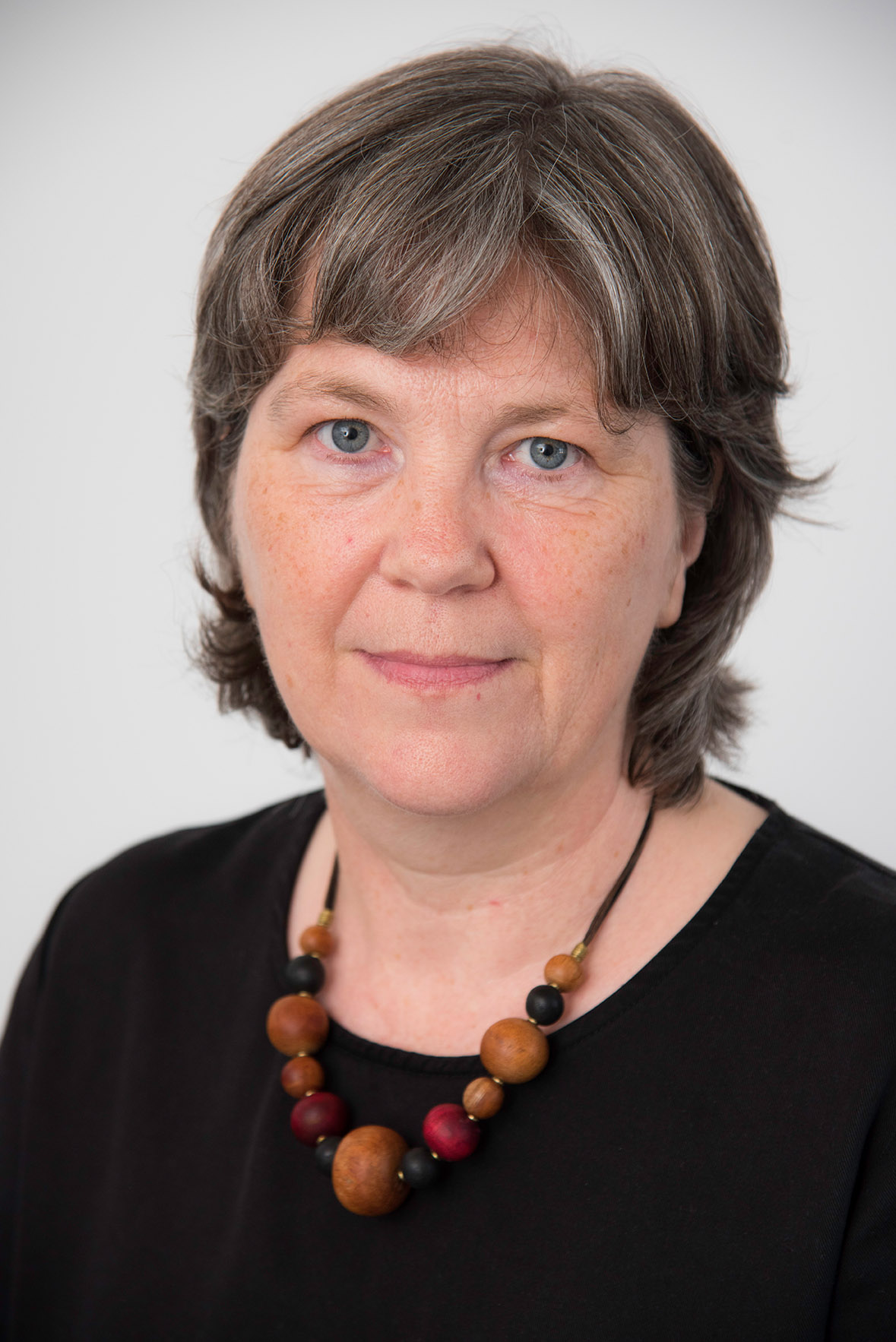}}\vspace*{100pt}}%
\end{wrapfigure}
\noindent\small 
{\bf Monique Snoeck} is full professor at the KU Leuven, Research Center of Information Systems Engineering (LIRIS), and visiting professor at the UNamur. Her research focuses on conceptual modelling, enterprise modeling, requirements engineering, model-driven engineering and business process management, the teaching and learning of these topics and learning analytics. Previous research has resulted in the Enterprise Information Systems Engineering approach MERODE, and its companion e-learning and prototyping tools MERLIN and JMermaid. In the domains of Smart Learning environments and Technology enhanced learning she focuses on intelligent feedback provisioning and learning analytics with the aim of improving learning environments so as to foster learner success and learner engagement. She has (co-)authored over 130 peer reviewed papers, half of which peer-reviewed journal papers. She is involved in numerous conferences in the domains of Information Systems such as CAiSE, PoEM, ER, EMMSAD, etc.\vadjust{\vspace{40pt}}}

\vbox{%
\begin{wrapfigure}{l}{80pt}
{\vspace*{-20pt}\fbox{\includegraphics[width=3cm]{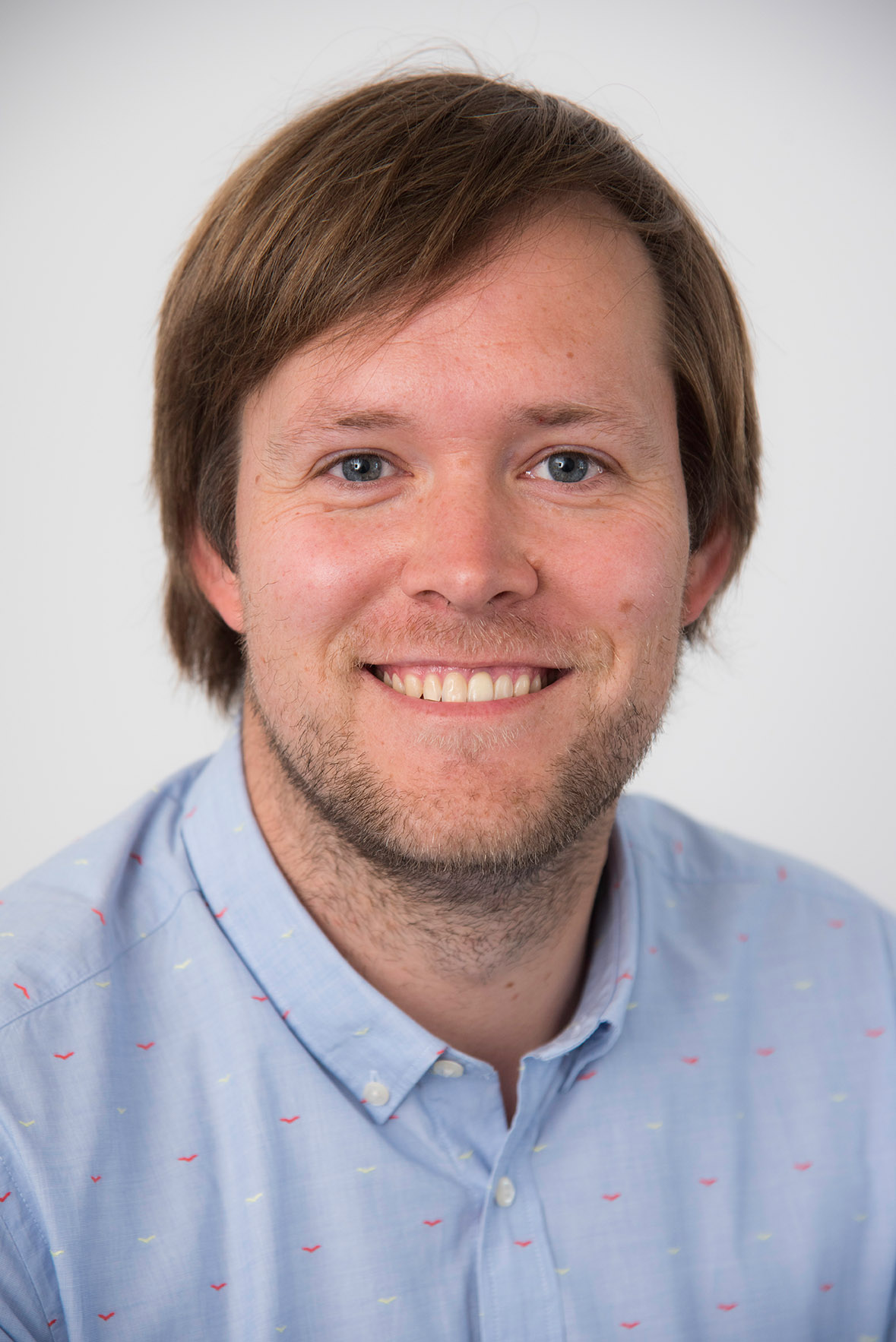}}\vspace*{100pt}}%
\end{wrapfigure}
\noindent\small 
{\bf Jochen De Weerdt} is an Associate Professor at the Research Centre of Information Systems Engineering (LIRIS) of the Faculty of Economics and Business at KU Leuven. He obtained a PhD in Business Economics at KU Leuven in 2012 on the topic of Automated Business Process Discovery. Subsequently, he worked as a postdoctoral research fellow at the Information Systems School of the Queensland University of Technology (Brisbane, Australia). His research and teaching interests include Business Information Systems, Business Analytics, Process Mining, Machine Learning, Learning Analytics, and Business Process Management. Jochen De Weerdt has published over 60 papers on these topics, including academic book chapters, peer-reviewed journal articles, and refereed papers at international conferences and workshops. He published papers in renowned journals such as IEEE Transactions on Knowledge and Data Engineering, Data Mining and Knowledge Discovery, Decision Support Systems, Information Systems, Computers in Human Behavior, etc. In addition, his findings have been presented at well-known international conferences such as BPM, ECML-PKDD, DSAA, and IEEE CEC. \vadjust{\vspace{40pt}}}

\end{document}